\documentclass[conference]{IEEEtran}
\IEEEoverridecommandlockouts
\usepackage{cite}
\usepackage{amsmath,amssymb,amsfonts}
\usepackage{algorithmic}
\usepackage{graphicx}
\usepackage{textcomp}
\usepackage{xcolor}

\usepackage{hyperref}
\usepackage{bm}
\usepackage{subfigure}
\usepackage{booktabs}
\usepackage{multirow}
\usepackage{arydshln}
\usepackage{colortbl}

\makeatletter
\def\adl@drawiv#1#2#3{%
        \hskip.0\tabcolsep
        \xleaders#3{#2.5\@tempdimb #1{1}#2.5\@tempdimb}%
                #2\z@ plus1fil minus1fil\relax
        \hskip.0\tabcolsep}
\newcommand{\cdashlinelr}[1]{%
  \noalign{\vskip\aboverulesep
           \global\let\@dashdrawstore\adl@draw
           \global\let\adl@draw\adl@drawiv}
  \cdashline{#1}
  \noalign{\global\let\adl@draw\@dashdrawstore
           \vskip\belowrulesep}}
\makeatother

\begin{document}

\title{Perception-Oriented Latent Coding for High-Performance Compressed Domain\\Semantic Inference}

\author{
   Xu Zhang$^1$ \quad \quad
   Ming Lu$^1$\thanks{This work was supported in part by Natural Science Foundation of Jiangsu Province (Grant No. BK20241226) and Natural Science Foundation of China (Grant No. 62401251, 62431011). The authors would like to express their sincere gratitude to the Interdisciplinary Research Center for Future Intelligent Chips (Chip-X) and Yachen Foundation for their invaluable support.}\thanks{Correspondence to: Ming Lu \textless{}minglu@nju.edu.cn\textgreater{}.} \quad \quad
   Yan Chen$^2$ \quad \quad
   Zhan Ma$^1$ \\
   $^1$Nanjing University \quad $^2$Jiangsu Academy of Safety Science and Technology \\
   {\tt\small xu.zhang@smail.nju.edu.cn, \{minglu,mazhan\}@nju.edu.cn, ggs@jsajy.cn}
}

\maketitle

\begin{abstract}
In recent years, compressed domain semantic inference has primarily relied on learned image coding models optimized for mean squared error (MSE). However, MSE-oriented optimization tends to yield latent spaces with limited semantic richness, which hinders effective semantic inference in downstream tasks. Moreover, achieving high performance with these models often requires fine-tuning the entire vision model, which is computationally intensive, especially for large models. To address these problems, we introduce \textit{Perception-Oriented Latent Coding (POLC)}, an approach that enriches the semantic content of latent features for high-performance compressed domain semantic inference. With the semantically rich latent space, POLC requires only a plug-and-play adapter for fine-tuning, significantly reducing the parameter count compared to previous MSE-oriented methods. Experimental results demonstrate that POLC achieves rate-perception performance comparable to state-of-the-art generative image coding methods while markedly enhancing performance in vision tasks, with minimal fine-tuning overhead. Code is available at \url{https://github.com/NJUVISION/POLC}.
\end{abstract}

\begin{IEEEkeywords}
learned image coding, compressed domain semantic inference, perception-oriented optimization, compressed representation, deep learning
\end{IEEEkeywords}

\section{Introduction}
\label{sec:intro}

Image coding is fundamental for efficient visual data storage and transmission, playing a critical role in various applications, including multimedia streaming, autonomous systems, and remote intelligent analysis tasks. Traditional image coding methods, such as JPEG~\cite{wallace1991jpeg}, BPG~\cite{BPG}, and VVC Intra~\cite{VVC}, have been extensively used due to their effectiveness in preserving visual quality under compression. However, their reliance on heuristic-driven algorithms limits their adaptability to the complex and varied demands of machine vision applications. The emergence of learned image coding (LIC)~\cite{balle2017endtoend, balle2018variational, minnen2018joint, lu2022transformerbased, lu2022highefficiency, he2022elic, liu2023learned, mentzer2020highfidelity, muckley2023improving} has revolutionized the field. By leveraging end-to-end data-driven optimization, LIC models have demonstrated impressive improvements in rate-distortion (R-D) and rate-perception (R-P) performance, learning efficient and flexible representations. This adaptability has extended LIC's application to vision tasks beyond human-centric perception~\cite{song2021variablerate, chen2023transtic, li2024image, zhang2024allinone}.

\begin{figure}[t]\centering
\subfigure{\includegraphics[scale=0.463]{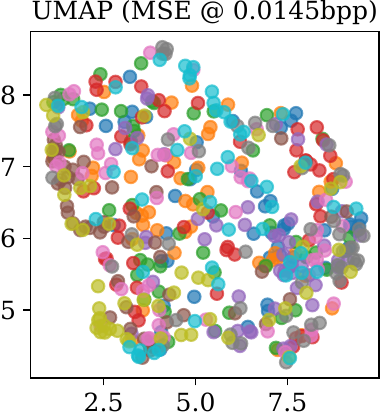}}\label{subfig:umap_mse}\hspace{2pt}
\subfigure{\includegraphics[scale=0.463]{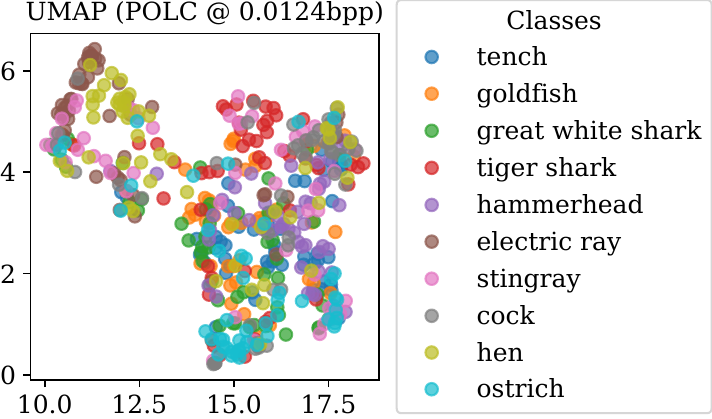}}\label{subfig:umap_polc}
\vspace{-0.3\baselineskip}
\caption{Latent space visualization of MSE- and perception-oriented optimization using UMAP~\cite{mcinnes2018umap} on ImageNet-1K~\cite{imagenet1k}. Compared to MSE optimization, POLC provides a more discriminative latent space, where data points of the same class are closer together, serving as a better initialization for fine-tuning and enabling higher performance with significantly fewer training parameters.
}
\label{fig:umap}
\vspace{-1\baselineskip}
\end{figure}

Beyond reconstruction, LIC has been extended to support compressed domain semantic inference, enabling latent representations generated during compression to directly serve as input for downstream vision tasks~\cite{torfason2018towards, liu2022improving, duan2023unified}. This approach eliminates the need for fully decoding images, offering potential efficiency gains. However, current methods predominantly reply on LIC models optimized for mean squared error (MSE), focusing on pixel-level reconstruction accuracy while neglecting the semantic richness and discriminability of latent space (see Fig.~\ref{fig:umap}) which are essential for effective semantic inference in complex tasks. Additionally, achieving high performance with these models typically requires fine-tuning the entire vision model, especially when adapting to new tasks. While effective, this approach is computationally expensive and impractical for large-scale vision models.

To overcome these limitations, we introduce \textbf{\textit{Perception-Oriented Latent Coding (POLC)}}, an approach that enhances the semantic richness of latent representations through perception-oriented optimization for high-performance inference. By training LIC models to prioritize semantic-level perceptual features, POLC improves performance across different downstream vision tasks and models. Moreover, in contrast to previous methods, POLC can achieve high performance with minimal fine-tuning only a universal plug-and-play adapter, thus significantly reducing training overhead. This approach not only bridges the gap between compression and vision tasks, but also introduces a training-friendly and computationally efficient framework for semantic inference.

Our contributions are summarized as follows:
\begin{itemize}
    \item We investigate perception-oriented optimization for latent coding that enhances the semantic richness of latent features, enabling effective compressed domain semantic inference without compromising reconstruction quality.
    \item By leveraging semantically enriched latent representations, POLC reduces the reliance on task-specific vision model fine-tuning, requiring only a plug-and-play adapter for high performance in downstream tasks.
    \item We conduct comprehensive evaluations to demonstrate that POLC achieves R-P performance comparable to state-of-the-art (SOTA) generative image coding methods while markedly improving downstream vision task performance with minimal fine-tuning overhead.
\end{itemize}

\section{Related Work}
\label{sec:related}

Semantic inference in coding scenarios has become a growing focus within LIC research, driven by the increasing need to support downstream vision tasks efficiently. The extension of LIC to semantic inference typically focuses on two paradigms: handling tasks using reconstructed images and performing task inference directly in the compressed domain.

The first paradigm involves handling machine vision tasks using reconstructed images, where the image is fully decoded before task-specific analysis. This approach typically involves distinct encoder-decoder pairs for task-specific optimization, but multiple models and bitstreams introduce significant parameter and bitrate overhead~\cite{song2021variablerate, chen2023transtic, li2024image}. To mitigate these issues, Zhang et al.~\cite{zhang2024allinone} proposed multi-path aggregation (MPA), which allocates latent features among task-specific paths based on their importance to different tasks within a unified model and representation. While MPA yields high performance, it still requires decoding the full image for high-performance semantic inference, leading to additional latency and computational overhead.

The second paradigm, performing analysis directly in the compressed domain, has gained attention for its potential to bypass full image reconstruction, thereby reducing latency and computational overhead. This approach utilizes compressed latent representations as inputs for vision tasks, enabling faster and more efficient inference~\cite{torfason2018towards, liu2022improving, feng2022omnipotent, duan2023unified}. Liu et al.~\cite{liu2022improving} implemented gate modules to select the most important channels for each task. Feng et al.~\cite{feng2022omnipotent} compressed intermediate features from a vision backbone to create generic representations suitable for various tasks. Duan et al.~\cite{duan2023unified} introduced adapters to bridge compressed representations with task-specific vision backbones, enabling direct analysis. Additionally, scalable coding techniques~\cite{liu2021semantics, yan2021sssic, choi2022scalable} embed multiple nested bitstreams to support various tasks. However, managing layered representations without redundancy remains a challenge. While these approaches improve efficiency by skipping image reconstruction, they often fail to fully exploit the semantic richness of latent representations, limiting performance in vision tasks.

Despite the advancements in these paradigms, current approaches face limitations in balancing semantic richness, fine-tuning efficiency, and task performance. This highlights the need for methods that can directly enhance the semantic content of compressed latent representations while supporting efficient and high-performance inference.

\section{Towards High-Performance Inference}
\label{sec:method}

Achieving high-performance inference requires overcoming the limitations of latent space in terms of semantic richness and fine-tuning efficiency. To this end, we explore the performance differences in semantic inference between the introduced \textit{POLC} and MSE-optimized models, demonstrating how POLC enhances the semantic richness of latent representations to improve inference capabilities. Additionally, we design a \textit{Universal Adapter} that seamlessly bridges image coders and modern vision models while minimizing the training overhead.

\begin{figure}[t]\centering
\includegraphics[scale=0.4]{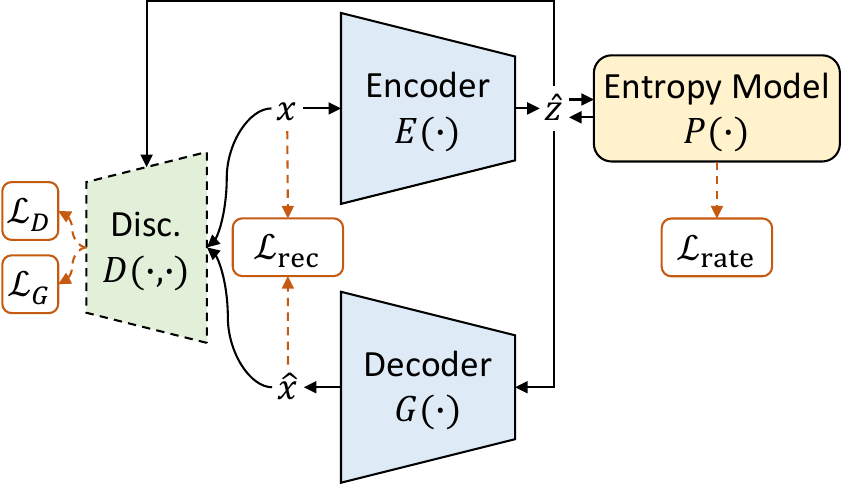}
\vspace{-0.5\baselineskip}
\caption{Perception-oriented latent coding. It leverages the generative image coding framework~\cite{mentzer2020highfidelity} by incorporating discriminator loss $\mathcal{L}_D$, generator loss $\mathcal{L}_G$, reconstruction loss $\mathcal{L}_\text{rec}$ and bitrate loss $\mathcal{L}_\text{rate}$, making the latents $\hat{\bm{z}}$ semantically richer, which in turn improves the performance of inference.}
\label{fig:polc}
\vspace{-0.5\baselineskip}
\end{figure}

\subsection{Perception-Oriented Latent Coding}
\label{sec:polc}

Existing LIC models, predominantly optimized for MSE, often prioritize reconstruction fidelity over semantic richness. To enable high-performance semantic inference, the latent space produced by the encoder $E(\cdot)$ must capture semantic-level features beyond pixel-level differences. To address this imbalance, POLC shifts the optimization focus of latent coding from traditional MSE-based objectives to a perception-oriented approach, as shown in Fig.~\ref{fig:polc}. Unlike MSE optimization, which emphasizes pixel-level reconstruction accuracy, POLC aims to capture perceptual features critical for downstream vision tasks with Generative Adversarial Network (GAN)~\cite{ian2014generative} while ensuring competitive reconstruction quality.

Specifically, the optimization objective incorporates a perceptual loss term, $\mathcal{L}_\text{perc}$, in reconstruction loss $\mathcal{L}_\text{rec}$ to align latent features with semantic information as demonstrated in~\cite{zhang2024allinone}. Furthermore, to ensure visually appealing reconstruction by the decoder $G(\cdot)$, the coder is trained under the supervision of a conditional discriminator $D(\text{cond.}, \text{input})$, following the same practices in generative image coding~\cite{mentzer2020highfidelity}:
\begin{align}
    \mathcal{L}_D &= \mathbb{E}_{\bm{\hat z} \sim p_{\bm{z}}}[-\log (1-D(\bm{\hat z}, G(\bm{\hat z}))] \nonumber \\
    &~\quad+ \mathbb{E}_{\bm{x}\sim p_{\bm{x}} }[-\log D(E(\bm{x}), \bm{x})],\label{eq:loss_d} \\
    \mathcal{L}_G &= \mathbb{E}_{\bm{\hat z} \sim p_{\bm{z}}}[-\log ( D(\bm{\hat z}, G(\bm{\hat z})) )],\label{eq:loss_g} \\
    \mathcal{L}_\text{rec} &= \mathbb{E}_{\bm{x}\sim p_{\bm{x}}} [\lambda_{d} d(\bm{x}, \hat{\bm{x}}) + \lambda_\text{perc} \mathcal{L}_\text{perc}],\label{eq:loss_rec} \\
    \mathcal{L}_{EGP} &= \lambda_\text{rate} \mathcal{L}_\text{rate} + \lambda_\text{rec} \mathcal{L}_\text{rec} + \lambda_{G} \mathcal{L}_{G}, \label{eq:loss_egp}
\end{align}
where $\bm{x}$, $\bm{\hat{x}}$, $\bm{z}$, and $\bm{\hat{z}}$ represent the input image, reconstructed image, and compressed latents before and after quantization, respectively. $\mathcal{L}_\text{rate}$ denotes the bitrate $\mathbb{E}_{\bm{x} \sim p_{\bm x}}[-\log_{2}p_{\bm{\hat{z}}}(\bm{\hat{z}})]$ estimated by the entropy model $P(\cdot)$, and $d(\cdot, \cdot)$ corresponds to distortion $\text{MSE}(\cdot, \cdot)$.
To enhance the semantic expressiveness of the latent space, we adopt Learned Perceptual Image Patch Similarity (LPIPS)~\cite{zhang2018unreasonable} between $\bm{x}$ and $\bm{\hat{x}}$ as $\mathcal{L}_{\text{perc}}$, effectively aligning latent features with semantic-level perception.

Through perception-oriented optimization, POLC achieves two critical objectives. First, it aligns with existing generative image coding methods to ensure high-quality image reconstruction. Second, it enhances the semantic richness of the latent space, significantly improving inference performance and enabling efficient training of downstream tasks.

\begin{figure}[t]\centering
\includegraphics[scale=0.4]{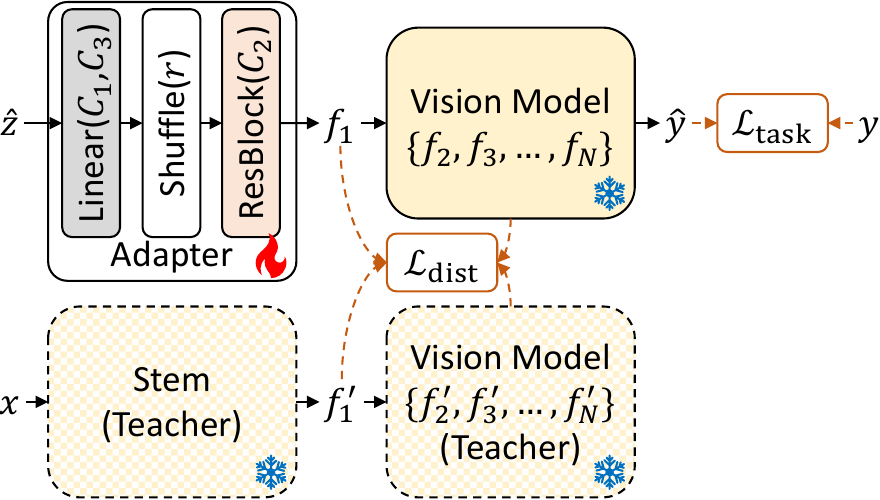}
\caption{Universal adapter. With POLC, only the adapter need to be trained by minimizing the task loss $\mathcal{L}_\text{task}$ between predicted $\hat{\bm{y}}$ and ground truth $\bm{y}$, along with the distillation loss $\mathcal{L}_\text{dist}$ between features $\bm{f}_i$ and $\bm{f}'_i$. Note that the dashed boxes for knowledge transfer~\cite{duan2023unified} will be discarded after training.}
\label{fig:adapter}
\vspace{-0.5\baselineskip}
\end{figure}

\subsection{Universal Adapter for Modern Models}
\label{sec:adapter}

\paragraph{Design}
The design of vision models has undergone significant evolution with the advancement of computer vision. Particularly, the structure of the initial feature extraction layers, commonly referred to as the \textit{stem}, has diversified across traditional and modern architectures. In conventional convolutional neural networks (CNNs) such as ResNet~\cite{he2016deep}, the stem typically consists of overlapped convolutions followed by normalization and activation, performing an initial downsampling by $2\times$, which is then followed by max pooling to achieve a total downsampling of $4\times$. In contrast, modern hierarchical models like ConvNeXt~\cite{liu2022convnet} and Swin Transformer~\cite{liu2021swin} adopt \textit{Patch Embedding} as their stem, using a single-layer unoverlapped convolution to directly downsample by $4\times$. Furthermore, isotropic models such as Vision Transformer (ViT)~\cite{dosovitskiy2021vit, touvron2021deit} employ an even more aggressive approach, directly downsampling by $16\times$ in the stem. Given the increasing diversity in stem designs across traditional and modern vision models, adaptation methods that target specific stem structures such as~\cite{torfason2018towards, duan2023unified} become less generalizable and require modifications to accommodate each stem's unique architecture. This lack of generality limits their applicability when dealing with heterogeneous model structures.

To address this limitation, we propose a \textit{Universal Adapter} that bypasses the stem altogether by directly performing spatial and channel dimensional mapping. Instead of adapting to specific stem designs, the universal adapter focuses solely on aligning the channel count and spatial resolution of the output features to the requirements of the downstream vision model, as illustrated in Fig.~\ref{fig:adapter}. Our adapter design prioritizes simplicity and generalizability, comprising only upsampling and a ResBlock. By decoupling the adaptation process from the stem's structural variations, the adapter provides a unified and efficient solution compatible with a wide range of vision model architectures. Specifically, given input latent $\hat{\bm{z}} \in \mathbb{R}^{H_1 \times W_1 \times C_1}$ and target feature $\bm{f}'_1 \in \mathbb{R}^{H_2 \times W_2 \times C_2}$, the adapter consists of three key components:
\begin{itemize}
    \item \textbf{Linear Channel Projection:} A fully connected layer is used to perform a linear projection of the channel dimensions, aligning the channel count of the latent features with the requirements of the following pixel shuffle and the downstream vision model. The input has $C_1$ channels and the output has $C_3 = r^2 C_2$ channels.
    \item \textbf{Pixel Shuffle for Spatial Alignment:} A pixel shuffle layer is employed to adjust the spatial resolution of the latent features to match the input resolution expected by the downstream vision model. The upsampling factor is set to $r = \frac{H_2 \times W_2}{H_1 \times W_1}$.
    \item \textbf{Residual Mapping:} A residual block identical to those used in the LIC model is incorporated to perform a learnable transformation of the latent features, enhancing the alignment between $\bm{f}_i$ and $\bm{f}'_i$. The number of channels for both input and output features is $C_2$.
\end{itemize}
This design enables the adapter to provide a consistent interface for adapting features to a wide range of vision models, regardless of their specific stem architectures.

\paragraph{Training Strategy}
POLC significantly reduces the training burden by producing semantically enriched latent features that are directly compatible with downstream tasks. This design allows the adapter to be efficiently fine-tuned without requiring joint training of the entire vision model. As a result, even for large-scale vision models, the training overhead remains minimal, making the framework scalable.

During training, the whole LIC model in Fig.~\ref{fig:polc} is kept frozen to ensure that the quality of the reconstructed images is not affected. The objective loss $\mathcal{L}_\text{adapt}$ is formed as:
\begin{align}
\mathcal{L}_\text{task} &= \text{Task-Criterion}(\bm{y}, \hat{\bm{y}}), \\
\mathcal{L}_\text{dist} &= {\textstyle \sum_{i=1}^{N}{\lambda_i d(\bm{f}'_i, \bm{f}_i)}}, \\
\mathcal{L}_\text{adapt} &= \lambda_\text{rate} \mathcal{L}_\text{rate} + \lambda_\text{task} \mathcal{L}_\text{task} + \lambda_\text{dist} \mathcal{L}_\text{dist},
\end{align}
where $\mathcal{L}_\text{task}$ represents the task-specific loss (e.g., cross-entropy for classification) between the prediction $\hat{\bm{y}}$ and the ground truth $\bm{y}$, and $\mathcal{L}_\text{dist}$ is a distillation loss used to transfer knowledge~\cite{duan2023unified} from the original vision model trained on uncompressed images. $\mathcal{L}_\text{dist}$ ensures that the features $\bm{f}_i$ extracted by the compressed domain vision model at each stage, marked as solid boxes in Fig.~\ref{fig:adapter}, closely match those $\bm{f}'_i$ extracted by the teacher model marked as dashed boxes (will be discarded after training). We set $\lambda_1 = \lambda_2 = \dots = 1$ following~\cite{duan2023unified}. By decoupling the adapter's optimization from the full vision model, the framework achieves minimal computational overhead with high-performance inference, ensuring efficient adaptation to a wide range of vision tasks and models. Note that only the solid boxes in Fig.~\ref{fig:adapter} will be used during inference.

\begin{figure*}[t]\centering
\subfigure{\includegraphics[scale=0.461]{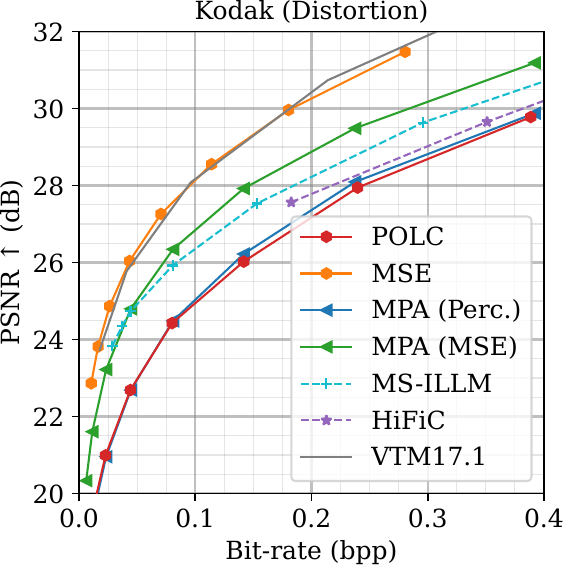}\label{subfig:kodak_rd}}
\subfigure{\includegraphics[scale=0.461]{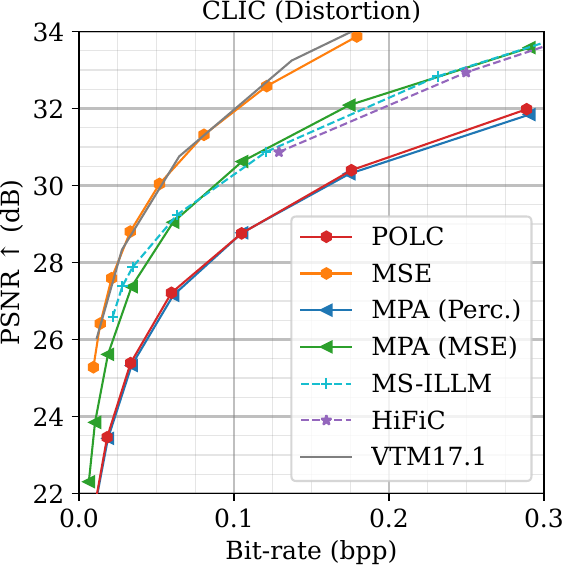}\label{subfig:clic_rd}}
\subfigure{\includegraphics[scale=0.461]{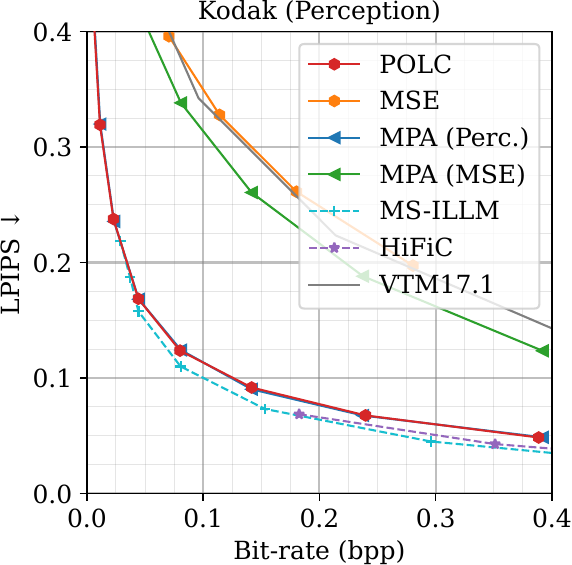}\label{subfig:kodak_lpips}}
\subfigure{\includegraphics[scale=0.461]{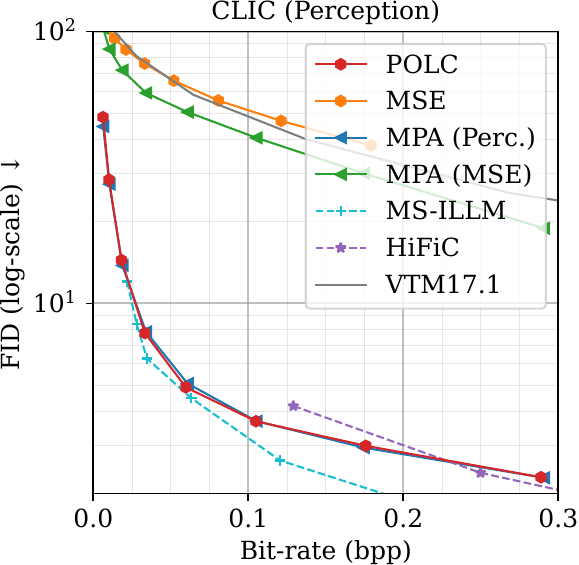}\label{subfig:clic_fid}}\vspace{-6pt}\\
\subfigure{\includegraphics[scale=0.464]{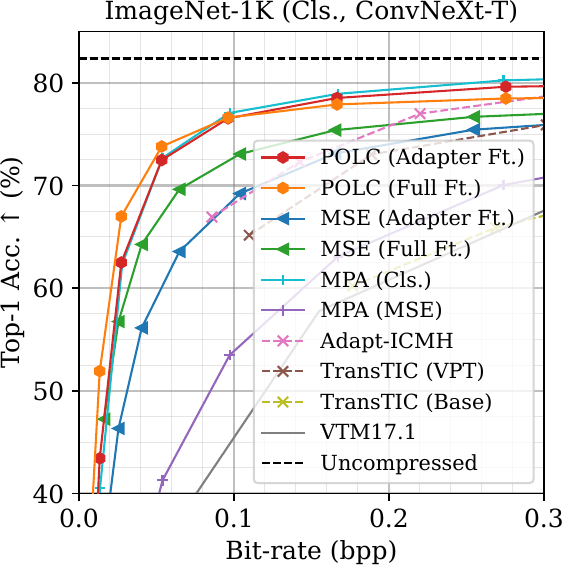}\label{subfig:convnext_t}}
\subfigure{\includegraphics[scale=0.464]{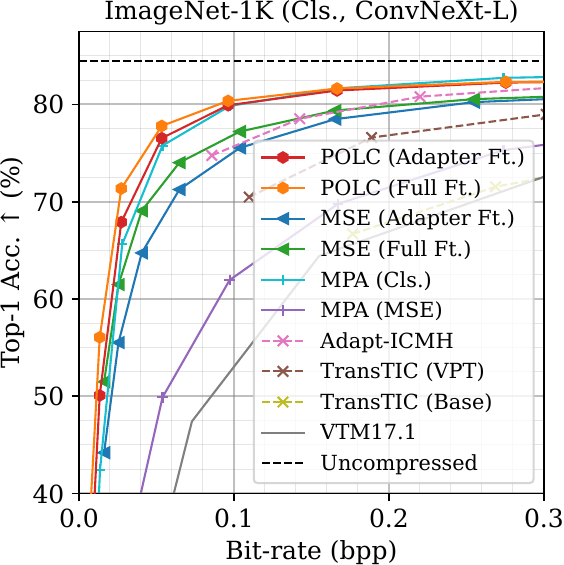}\label{subfig:convnext_l}}
\subfigure{\includegraphics[scale=0.464]{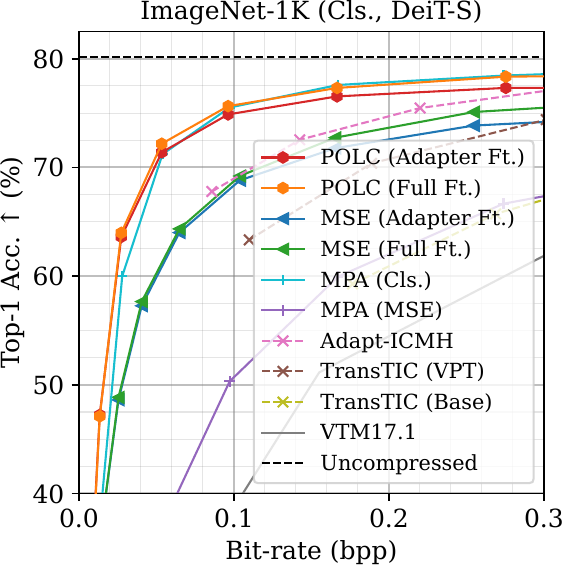}\label{subfig:deit_s}}
\subfigure{\includegraphics[scale=0.464]{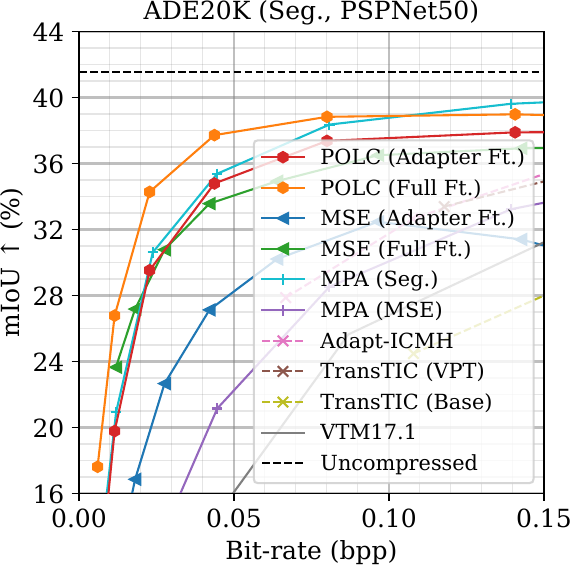}\label{subfig:pspnet50}}
\vspace{-1.5\baselineskip}
\caption{Reconstruction quality on the Kodak~\cite{kodak} and CLIC test set~\cite{clic2020} and vision task performance on ImageNet-1K~\cite{imagenet1k} and ADE20K~\cite{ade20k}. While achieving comparable performance to other generative image coding models, POLC supports high-performance compressed domain semantic inference across different vision tasks and different types/sizes of vision models. Adapter Ft. and Full Ft. denote fine-tuning the adapter and the whole vision model, respectively.}
\label{fig:performance}
\vspace{-1\baselineskip}
\end{figure*}

\section{Experiments}
\label{sec:experiments}

\subsection{Experimental Setup}
\label{sec:setup}

\paragraph{Datasets}
When conduct POLC training, a combined dataset is used including Flicker2W~\cite{flicker2w}, DIV2K~\cite{div2k}, and CLIC~\cite{clic2020}, about 23K images in total. For fine-tuning on downstream tasks, we use ImageNet-1K~\cite{imagenet1k} for classification and ADE20K~\cite{ade20k} for semantic segmentation.

\paragraph{Baselines}
To demonstrate the advantages of POLC, we compare against the following baselines:
\begin{itemize}
    \item \textbf{Pixel-Domain Semantic Inference:} We compare POLC with recent SOTA baselines, including TinyLIC-based~\cite{lu2022highefficiency} MPA~\cite{zhang2024allinone}, and TIC-based~\cite{lu2022transformerbased} Adapt-ICMH~\cite{li2024image} and TransTIC~\cite{chen2023transtic}. We also add VTM 17.1~\cite{VTM_17_1} intra profile as a baseline for the traditional image coding methods.
    \item \textbf{Compressed-Domain Semantic Inference:} Following the setup of~\cite{torfason2018towards, duan2023unified}, we evaluate models optimized with MSE to highlight the performance improvement brought by POLC training.
\end{itemize}

To further demonstrate the generalizability of POLC across different vision tasks and architectures, we evaluate both classification task and semantic segmentation task, following the setups of~\cite{duan2023unified, zhang2024allinone}. The chosen models include ConvNeXt~\cite{liu2022convnet}, DeiT~\cite{touvron2021deit}, and ResNet-based~\cite{he2016deep} PSPNet~\cite{zhao2017pyramid}, representing hierarchical, isotropic, and specialized segmentation architectures, respectively. Our models are implemented based on TinyLIC~\cite{lu2022highefficiency} with the variable-rate settings aligned with MPA~\cite{zhang2024allinone} for a fair comparison.

\paragraph{Training}
During POLC training, $\lambda_\text{rate}$ is randomly sampled from $\{18.0, 9.32, 4.83, 2.5, 1.3, 0.67, 0.35, 0.18\}$. The following loss coefficients are used: $\lambda_d = 1$, $\lambda_\text{perc} = 1$, $\lambda_\text{rec} = 1$, $\lambda_\text{task} = 1$ and $\lambda_G = 0.8$. $\lambda_\text{dist}$ should be adjusted according to the amplitude of $\mathcal{L}_\text{dist}$. We use $\lambda_\text{dist} = 0.001$ for ConvNeXt-L, $\lambda_\text{dist} = 0.01$ for ConvNeXt-T, $\lambda_\text{dist} = 0.1$ for DeiT-S, and $\lambda_\text{dist} = 10$ for PSPNet50. The data augmentation and training process for POLC follow MPA~\cite{zhang2024allinone}, with 3M steps for image coder training and 500K steps for vision task fine-tuning. For both stages, the initial learning rate is set to $10^{-4}$ and decayed to $10^{-5}$ for the final 25\% of steps. Adam~\cite{ADAM} is used for optimization and the batch size is set to 8 for all tasks. Notably, when fine-tuning for semantic segmentation, we use 512$\times$512 image patches since the training objective is task accuracy rather than on reconstruction fidelity like MPA~\cite{zhang2024allinone}.

\paragraph{Evaluation}
We evaluate image reconstruction quality using Peak Signal-to-Noise Ratio (PSNR), Fréchet Inception Distance (FID) and LPIPS to  assess R-D and R-P performance. For classification and semantic segmentation tasks, we report Top-1 Accuracy and mean Intersection over Union (mIoU), respectively. All evaluations follow the standard protocols established in prior work~\cite{mentzer2020highfidelity, muckley2023improving, zhang2024allinone} to ensure comparability.

\subsection{Main Results}

As shown in Fig.~\ref{fig:performance}, POLC demonstrates reconstruction quality comparable to SOTA generative image coding models such as HiFiC~\cite{mentzer2020highfidelity} and MS-ILLM~\cite{muckley2023improving}, validating its ability to capture rich semantic features essential for high-quality image reconstruction. Furthermore, extensive testing across various vision tasks and models reveals that, by fine-tuning only the adapter, POLC outperforms fully fine-tuned models with MSE-optimized LIC which represent previous methods~\cite{torfason2018towards, duan2023unified}, achieving performance similar to the SOTA pixel-domain semantic inference method MPA~\cite{zhang2024allinone}.

In particular, for classification tasks, fine-tuning only the adapter achieves performance similar to that of fully fine-tuned models, indicating that for global understanding tasks, fewer fine-tuning parameters are required. For semantic segmentation tasks, fully fine-tuning the model leads to further improvements, suggesting that increasing the number of fine-tuning parameters can embed more semantic information in individual latent features and yield optimal performance, which is beneficial to dense prediction tasks.

POLC and the proposed adapter also show great versatility across different vision models. By using POLC and replacing the original stem with the proposed adapter, significant performance improvements are achieved across various models, including ConvNeXt~\cite{liu2022convnet}, DeiT~\cite{touvron2021deit}, and ResNet-based~\cite{he2016deep} PSPNet~\cite{zhao2017pyramid}, demonstrating the generalizability of POLC and the adapter. Moreover, as the model size increases from ConvNeXt-T to ConvNeXt-L, performance continues to improve, highlighting the scalability for larger models.

\begin{figure}[t]\centering
{
\resizebox{\linewidth}{!}{
\begin{tabular}{@{\hskip 0mm}l@{\hskip 1.0mm}l@{\hskip 1.0mm}l@{\hskip 1.0mm}l@{\hskip 1.0mm}l@{\hskip 0mm}}

\toprule
Original &
Rec. (POLC) &
Bit Alloc. (POLC) &
Rec. (MSE) &
Bit Alloc. (MSE)
\\ \midrule

\texttt{val00035565}
& \multicolumn{2}{@{}l}{22.16dB PSNR, 0.116 LPIPS, 0.030bpp} 
& \multicolumn{2}{@{}l}{26.00dB PSNR, 0.219 LPIPS, 0.032bpp} \\ 
\includegraphics[width=0.29\linewidth]{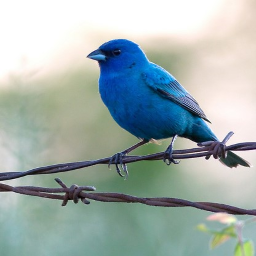} &
\includegraphics[width=0.29\linewidth]{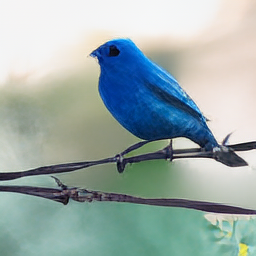} &
\includegraphics[width=0.358\linewidth]{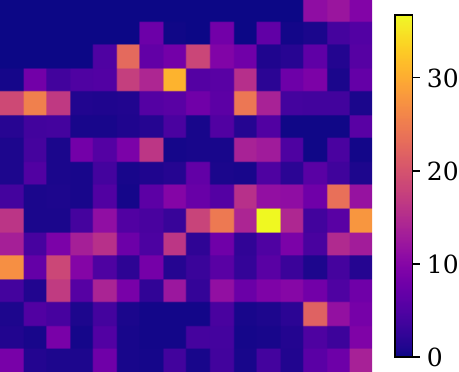} &
\includegraphics[width=0.29\linewidth]{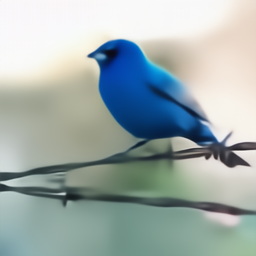} &
\includegraphics[width=0.358\linewidth]{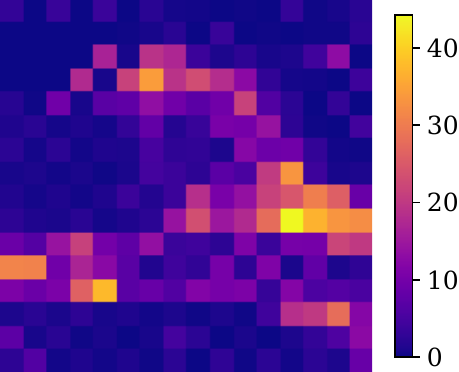} \\

\texttt{val00022926}
& \multicolumn{2}{@{}l}{20.41dB PSNR, 0.179 LPIPS, 0.032bpp} 
& \multicolumn{2}{@{}l}{23.94dB PSNR, 0.623 LPIPS, 0.034bpp} \\ 
\includegraphics[width=0.29\linewidth]{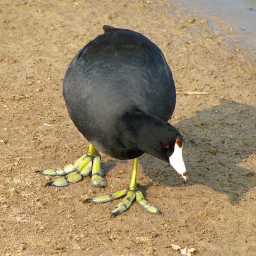} &
\includegraphics[width=0.29\linewidth]{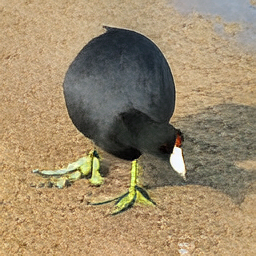} &
\includegraphics[width=0.358\linewidth]{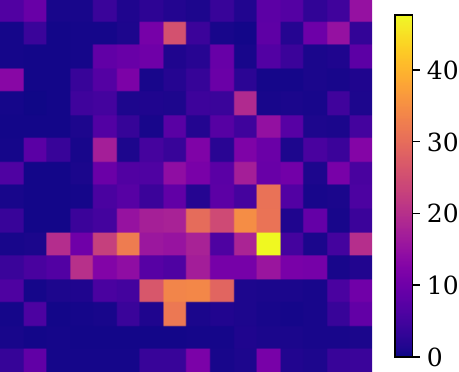} &
\includegraphics[width=0.29\linewidth]{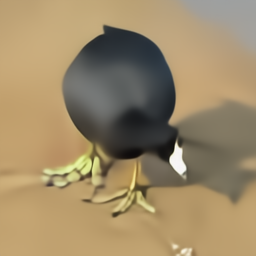} &
\includegraphics[width=0.358\linewidth]{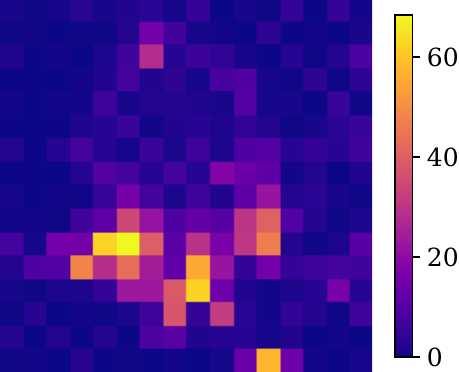} \\

\texttt{val00017577}
& \multicolumn{2}{@{}l}{20.35dB PSNR, 0.111 LPIPS, 0.039bpp} 
& \multicolumn{2}{@{}l}{23.17dB PSNR, 0.204 LPIPS, 0.036bpp} \\ 
\includegraphics[width=0.29\linewidth]{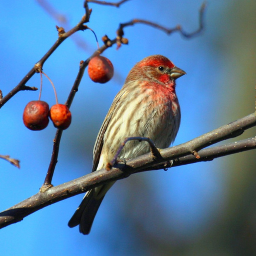} &
\includegraphics[width=0.29\linewidth]{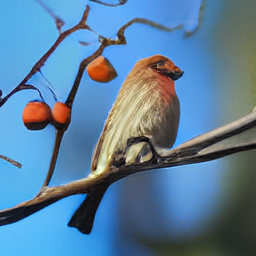} &
\includegraphics[width=0.358\linewidth]{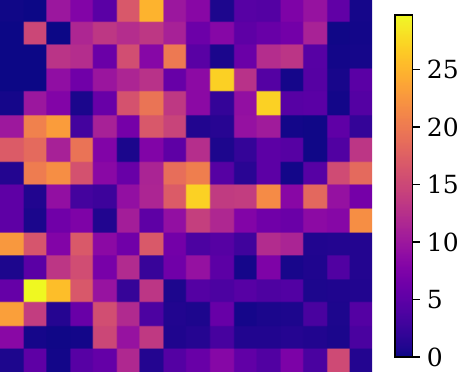} &
\includegraphics[width=0.29\linewidth]{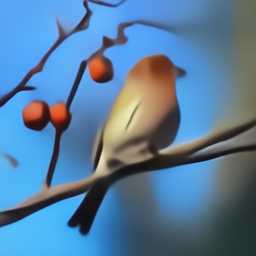} &
\includegraphics[width=0.358\linewidth]{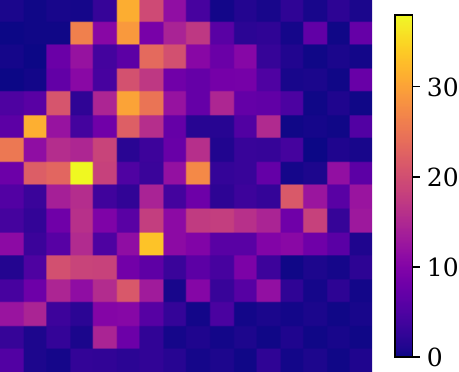} \\

\bottomrule
\end{tabular}
}
\vspace{-0.5\baselineskip}
}
\caption{Visualization of reconstruction and bit allocation. Compared to MSE optimization, POLC exhibits a more uniform bit allocation with smaller peaks in the high-frequency textures. This indicates that POLC focuses more on the distribution of semantic regions rather than just textures, highlighting its ability to prioritize semantic features for reconstruction and inference.}
\label{fig:vis}
\vspace{-\baselineskip}
\end{figure}

\subsection{Deep Dive}

To further investigate the differences between POLC and previous MSE-oriented optimization methods, we conduct a series of in-depth experiments and visualizations.

To study the differences in the latent space properties, we randomly sample 10 classes from the ImageNet validation set~\cite{imagenet1k}, and visualize the encoded latent representations $\hat{\bm{z}}$ using UMAP~\cite{mcinnes2018umap} shown in Fig.~\ref{fig:umap}. The results show that in the MSE-optimized latent space, data points are more dispersed, and samples from the same class fail to form effective clusters. This limits the performance of semantic inference and necessitates fine-tuning more parameters to adapt the vision model to the data distribution. In contrast, POLC’s latent space is more discriminative, with samples from the same class being closer to each other and partially forming clusters. This provides a better initialization for fine-tuning, enabling higher performance with fewer training parameters.

To further examine the semantic distribution of the learned representations, we visualize the reconstructed images and bit allocations in Fig.~\ref{fig:vis}. It can be observed that MSE-optimized models focus more on high-frequency textures to achieve higher PSNR, resulting in higher peaks in the bit allocation in these regions to faithfully reconstruct the fine details. On the other hand, POLC focuses more on the reconstruction of semantic objects, generating textures that are semantically similar but not exactly the same, thereby making the reconstructed image perceptually closer to the original. The bit allocation in POLC is more uniform and concentrated in the semantic regions. This demonstrates that POLC embeds more semantic information into the representations compared to MSE-optimized models, which is beneficial for vision tasks.

To quantitatively study the advantages of POLC, we perform a complexity analysis showcased in Table~\ref{tab:complexity}. The number of parameters that need to be fine-tuned during training is measured, as well as FLOPs and GPU latency on an NVIDIA RTX A6000 GPU during inference (including the entropy model, the adapter/decoder, and the vision model). As shown, POLC offers a significant advantage in terms of FLOPs and latency compared to MPA~\cite{zhang2024allinone}, while achieving similarly low fine-tuning parameter counts and comparable high accuracy. Compared to MSE-optimized models, POLC significantly reduces the fine-tuning parameter count and greatly improves inference performance. These results demonstrate the substantial advantages of POLC. We also conduct ablations in Table~\ref{tab:complexity} to analyze the impact of replacing pixel shuffle with transposed convolution (TConv) and removing ResBlocks. The results show that pixel shuffle and transposed convolution yield similar performance and complexity, whereas removing ResBlocks significantly degrades performance, validating that our design choices is both reasonable and effective.

\begin{table}[t]
\centering
\caption{Complexity analysis at a resolution of 512$\times$768 w.r.t. decoding and semantic inference. ConvNeXt-T~\cite{liu2022convnet} is used for inference, and the corresponding adapter costs 3.49GFLOPs.}
\vspace{-0.5\baselineskip}
\label{tab:complexity}
{
\resizebox{\linewidth}{!}{
\begin{tabular}{@{}lllll@{}}
\toprule
Methods & \#Params. for Ft.  & GFLOPs                 & Latency (ms)          & Acc. @ 0.1bpp \\ \midrule
MPA~\cite{zhang2024allinone} & 0.54M & 130.55    & 50.34                 & 77.06         \\ \cdashlinelr{1-5}
MSE     & 29.19M        & \multirow{2}{*}{55.12} & \multirow{2}{*}{9.76} & 73.06         \\
POLC    & 0.60M \textcolor[HTML]{ff0000}{(-28.59M)} & & & 76.54 \textcolor[HTML]{008000}{(+3.48)} \\
\hspace{1ex} Shuffle$\rightarrow$TConv & $+$0     & $+$0    & $+$0.04 & $+$0    \\
\hspace{1ex} w/o ResBlock & $-$0.11M & $-$2.74 & $-$1.64 & $-$0.79 \\ \bottomrule
\end{tabular}
}
}
\vspace{-\baselineskip}
\end{table}

\section{Conclusion}
\label{sec:conclusion}

In this paper, we introduce \textbf{\textit{Perception-Oriented Latent Coding (POLC)}} for high-performance compressed domain semantic inference. By leveraging generative image coding methods, POLC forms a discriminative latent space with rich semantic information. Merely fine-tuning a universal adapter that bridges image coders and vision models, POLC can achieve SOTA performance across differrent vision tasks and models. The main limitation of this approach is that the adapter needs to be modified and trained for each vision model. Future work will explore more generalizable methods for inference, aiming to reduce the need for model-specific adjustments.




\bibliographystyle{IEEEbib}
\bibliography{main}

\begin{thebibliography}{10}

\bibitem{wallace1991jpeg}
Gregory~K Wallace,
\newblock ``The {JPEG} still picture compression standard,''
\newblock {\em Communications of the ACM}, vol. 34, no. 4, pp. 30--44, 1991.

\bibitem{BPG}
Fabrice Bellard,
\newblock ``Bpg image format,'' 2014.

\bibitem{VVC}
ITU-T and ISO/IEC,
\newblock ``Versatile video coding,''
\newblock {\em ITU-T Rec. H.266 and ISO/IEC 23090-3}, 2020.

\bibitem{balle2017endtoend}
Johannes Ball{\'e}, Valero Laparra, and Eero~P. Simoncelli,
\newblock ``End-to-end optimized image compression,''
\newblock in {\em International Conference on Learning Representations}, 2017.

\bibitem{balle2018variational}
Johannes Ballé, David Minnen, Saurabh Singh, Sung~Jin Hwang, and Nick Johnston,
\newblock ``Variational image compression with a scale hyperprior,''
\newblock in {\em International Conference on Learning Representations}, 2018.

\bibitem{minnen2018joint}
David Minnen, Johannes Ball\'{e}, and George~D Toderici,
\newblock ``Joint autoregressive and hierarchical priors for learned image compression,''
\newblock in {\em Advances in Neural Information Processing Systems}. 2018, vol.~31, pp. 10794--10803, Curran Associates, Inc.

\bibitem{lu2022transformerbased}
Ming Lu, Peiyao Guo, Huiqing Shi, Chuntong Cao, and Zhan Ma,
\newblock ``Transformer-based image compression,''
\newblock in {\em 2022 Data Compression Conference (DCC)}, 2022, pp. 469--469.

\bibitem{lu2022highefficiency}
Ming Lu, Fangdong Chen, Shiliang Pu, and Zhan Ma,
\newblock ``High-efficiency lossy image coding through adaptive neighborhood information aggregation,''
\newblock {\em arXiv preprint arXiv:2204.11448}, 2022.

\bibitem{he2022elic}
Dailan He, Ziming Yang, Weikun Peng, Rui Ma, Hongwei Qin, and Yan Wang,
\newblock ``Elic: Efficient learned image compression with unevenly grouped space-channel contextual adaptive coding,''
\newblock in {\em Proceedings of the IEEE/CVF Conference on Computer Vision and Pattern Recognition (CVPR)}, June 2022, pp. 5718--5727.

\bibitem{liu2023learned}
Jinming Liu, Heming Sun, and Jiro Katto,
\newblock ``Learned image compression with mixed transformer-cnn architectures,''
\newblock in {\em Proceedings of the IEEE/CVF Conference on Computer Vision and Pattern Recognition (CVPR)}, June 2023, pp. 14388--14397.

\bibitem{mentzer2020highfidelity}
Fabian Mentzer, George~D Toderici, Michael Tschannen, and Eirikur Agustsson,
\newblock ``High-fidelity generative image compression,''
\newblock in {\em Advances in Neural Information Processing Systems}. 2020, vol.~33, pp. 11913--11924, Curran Associates, Inc.

\bibitem{muckley2023improving}
Matthew~J. Muckley, Alaaeldin El-Nouby, Karen Ullrich, Herve Jegou, and Jakob Verbeek,
\newblock ``Improving statistical fidelity for neural image compression with implicit local likelihood models,''
\newblock in {\em Proceedings of the 40th International Conference on Machine Learning}. 23--29 Jul 2023, vol. 202 of {\em Proceedings of Machine Learning Research}, pp. 25426--25443, PMLR.

\bibitem{song2021variablerate}
Myungseo Song, Jinyoung Choi, and Bohyung Han,
\newblock ``Variable-rate deep image compression through spatially-adaptive feature transform,''
\newblock in {\em Proceedings of the IEEE/CVF International Conference on Computer Vision (ICCV)}, October 2021, pp. 2380--2389.

\bibitem{chen2023transtic}
Yi-Hsin Chen, Ying-Chieh Weng, Chia-Hao Kao, Cheng Chien, Wei-Chen Chiu, and Wen-Hsiao Peng,
\newblock ``Transtic: Transferring transformer-based image compression from human perception to machine perception,''
\newblock in {\em Proceedings of the IEEE/CVF International Conference on Computer Vision (ICCV)}, October 2023, pp. 23297--23307.

\bibitem{li2024image}
Han Li, Shaohui Li, Shuangrui Ding, Wenrui Dai, Maida Cao, Chenglin Li, Junni Zou, and Hongkai Xiong,
\newblock ``Image compression for machine and human vision with spatial-frequency adaptation,''
\newblock in {\em Computer Vision -- ECCV 2024}, Cham, 2024, pp. 382--399, Springer Nature Switzerland.

\bibitem{zhang2024allinone}
Xu~Zhang, Peiyao Guo, Ming Lu, and Zhan Ma,
\newblock ``All-in-one image coding for joint human-machine vision with multi-path aggregation,''
\newblock in {\em Advances in Neural Information Processing Systems}. 2024, vol.~37, pp. 71465--71503, Curran Associates, Inc.

\bibitem{mcinnes2018umap}
Leland McInnes, John Healy, and James Melville,
\newblock ``Umap: Uniform manifold approximation and projection for dimension reduction,''
\newblock {\em arXiv preprint arXiv:1802.03426}, 2018.

\bibitem{imagenet1k}
Jia Deng, Wei Dong, Richard Socher, Li-Jia Li, Kai Li, and Li~Fei-Fei,
\newblock ``Imagenet: A large-scale hierarchical image database,''
\newblock in {\em 2009 IEEE Conference on Computer Vision and Pattern Recognition}, 2009, pp. 248--255.

\bibitem{torfason2018towards}
Róbert Torfason, Fabian Mentzer, Eiríkur Ágústsson, Michael Tschannen, Radu Timofte, and Luc~Van Gool,
\newblock ``Towards image understanding from deep compression without decoding,''
\newblock in {\em International Conference on Learning Representations}, 2018.

\bibitem{liu2022improving}
Jinming Liu, Heming Sun, and Jiro Katto,
\newblock ``Improving multiple machine vision tasks in the compressed domain,''
\newblock in {\em 2022 26th International Conference on Pattern Recognition (ICPR)}, 2022, pp. 331--337.

\bibitem{duan2023unified}
Zhihao Duan, Zhan Ma, and Fengqing Zhu,
\newblock ``Unified architecture adaptation for compressed domain semantic inference,''
\newblock {\em IEEE Trans. Circuit Syst. Video Technol.}, vol. 33, no. 8, pp. 4108--4121, 2023.

\bibitem{feng2022omnipotent}
Ruoyu Feng, Xin Jin, Zongyu Guo, Runsen Feng, Yixin Gao, Tianyu He, Zhizheng Zhang, Simeng Sun, and Zhibo Chen,
\newblock ``Image coding for machines with omnipotent feature learning,''
\newblock in {\em Computer Vision -- ECCV 2022}, Cham, 2022, pp. 510--528, Springer Nature Switzerland.

\bibitem{liu2021semantics}
Kang Liu, Dong Liu, Li~Li, Ning Yan, and Houqiang Li,
\newblock ``Semantics-to-signal scalable image compression with learned revertible representations,''
\newblock {\em International Journal of Computer Vision}, vol. 129, no. 9, pp. 2605--2621, 2021.

\bibitem{yan2021sssic}
Ning Yan, Changsheng Gao, Dong Liu, Houqiang Li, Li~Li, and Feng Wu,
\newblock ``Sssic: Semantics-to-signal scalable image coding with learned structural representations,''
\newblock {\em IEEE Transactions on Image Processing}, vol. 30, pp. 8939--8954, 2021.

\bibitem{choi2022scalable}
Hyomin Choi and Ivan~V. Bajić,
\newblock ``Scalable image coding for humans and machines,''
\newblock {\em IEEE Transactions on Image Processing}, vol. 31, pp. 2739--2754, 2022.

\bibitem{ian2014generative}
Ian Goodfellow, Jean Pouget-Abadie, Mehdi Mirza, Bing Xu, David Warde-Farley, Sherjil Ozair, Aaron Courville, and Yoshua Bengio,
\newblock ``Generative adversarial nets,''
\newblock in {\em Advances in Neural Information Processing Systems}. 2014, vol.~27, pp. 2672--2680, Curran Associates, Inc.

\bibitem{zhang2018unreasonable}
Richard Zhang, Phillip Isola, Alexei~A. Efros, Eli Shechtman, and Oliver Wang,
\newblock ``The unreasonable effectiveness of deep features as a perceptual metric,''
\newblock in {\em Proceedings of the IEEE Conference on Computer Vision and Pattern Recognition (CVPR)}, June 2018, pp. 586--595.

\bibitem{he2016deep}
Kaiming He, Xiangyu Zhang, Shaoqing Ren, and Jian Sun,
\newblock ``Deep residual learning for image recognition,''
\newblock in {\em Proceedings of the IEEE Conference on Computer Vision and Pattern Recognition (CVPR)}, June 2016, pp. 770--778.

\bibitem{liu2022convnet}
Zhuang Liu, Hanzi Mao, Chao-Yuan Wu, Christoph Feichtenhofer, Trevor Darrell, and Saining Xie,
\newblock ``A convnet for the 2020s,''
\newblock in {\em Proceedings of the IEEE/CVF Conference on Computer Vision and Pattern Recognition (CVPR)}, June 2022, pp. 11976--11986.

\bibitem{liu2021swin}
Ze~Liu, Yutong Lin, Yue Cao, Han Hu, Yixuan Wei, Zheng Zhang, Stephen Lin, and Baining Guo,
\newblock ``Swin transformer: Hierarchical vision transformer using shifted windows,''
\newblock in {\em Proceedings of the IEEE/CVF International Conference on Computer Vision (ICCV)}, October 2021, pp. 10012--10022.

\bibitem{dosovitskiy2021vit}
Alexey Dosovitskiy, Lucas Beyer, Alexander Kolesnikov, Dirk Weissenborn, Xiaohua Zhai, Thomas Unterthiner, Mostafa Dehghani, Matthias Minderer, Georg Heigold, Sylvain Gelly, Jakob Uszkoreit, and Neil Houlsby,
\newblock ``An image is worth 16x16 words: Transformers for image recognition at scale,''
\newblock in {\em International Conference on Learning Representations}, 2021.

\bibitem{touvron2021deit}
Hugo Touvron, Matthieu Cord, Matthijs Douze, Francisco Massa, Alexandre Sablayrolles, and Herve Jegou,
\newblock ``Training data-efficient image transformers \& distillation through attention,''
\newblock in {\em Proceedings of the 38th International Conference on Machine Learning}. 18--24 Jul 2021, vol. 139 of {\em Proceedings of Machine Learning Research}, pp. 10347--10357, PMLR.

\bibitem{kodak}
Eastman Kodak,
\newblock ``Kodak lossless true color image suite,'' 1993.

\bibitem{clic2020}
George Toderici, Wenzhe Shi, Radu Timofte, Lucas Theis, Johannes Balle, Eirikur Agustsson, Nick Johnston, and Fabian Mentzer,
\newblock ``Workshop and challenge on learned image compression (clic2020),'' 2020.

\bibitem{ade20k}
Bolei Zhou, Hang Zhao, Xavier Puig, Tete Xiao, Sanja Fidler, Adela Barriuso, and Antonio Torralba,
\newblock ``Semantic understanding of scenes through the ade20k dataset,''
\newblock {\em International Journal of Computer Vision}, vol. 127, pp. 302--321, 2019.

\bibitem{flicker2w}
Jiaheng Liu, Guo Lu, Zhihao Hu, and Dong Xu,
\newblock ``A unified end-to-end framework for efficient deep image compression,''
\newblock {\em arXiv preprint arXiv:2002.03370}, 2020.

\bibitem{div2k}
Eirikur Agustsson and Radu Timofte,
\newblock ``Ntire 2017 challenge on single image super-resolution: Dataset and study,''
\newblock in {\em Proceedings of the IEEE Conference on Computer Vision and Pattern Recognition (CVPR) Workshops}, July 2017.

\bibitem{VTM_17_1}
``Versatile video coding reference software version 17.1,'' \url{https://vcgit.hhi.fraunhofer.de/jvet/VVCSoftware_VTM/tags/VTM-17.1}, July 2022.

\bibitem{zhao2017pyramid}
Hengshuang Zhao, Jianping Shi, Xiaojuan Qi, Xiaogang Wang, and Jiaya Jia,
\newblock ``Pyramid scene parsing network,''
\newblock in {\em Proceedings of the IEEE Conference on Computer Vision and Pattern Recognition (CVPR)}, July 2017, pp. 2881--2890.

\bibitem{ADAM}
Diederik~P Kingma and Jimmy Ba,
\newblock ``Adam: A method for stochastic optimization,''
\newblock in {\em the 3rd Int. Conf. on Learning Representations}, 2015.

\end{thebibliography}

\end{document}